\documentclass[runningheads]{llncs}
\usepackage[T1]{fontenc}
\usepackage{graphicx}
\usepackage{amsmath}
\usepackage{amsfonts} % Provides the \mathbb command
\usepackage{cleveref} 
\usepackage{booktabs}
\usepackage{multirow, multicol}
\usepackage{todonotes}
\usepackage{xcolor}
\usepackage{amssymb}

\newcommand{\methodname}{ConsFormer}

\begin{document}

\title{Large Neighborhood Search meets \\ Iterative Neural Constraint Heuristics}

\author{Yudong W. Xu\inst{1}
% \orcidID{0009-0003-4601-4985} 
\and Wenhao Li\inst{1}
% \orcidID{1111-2222-3333-4444} 
\and Scott Sanner\inst{1,2}
% \orcidID{0000-0001-7984-8394} 
\and Elias B. Khalil\inst{1}
% \orcidID{0000-0001-5844-9642}
}
\authorrunning{YW. Xu et al.}
\institute{Department of Mechanical \& Industrial Engineering, University of Toronto \and
Vector Institute 
\\
\email{wil.xu@mail.utoronto.ca},
\email{chriswenhao.li@mail.utoronto.ca},
\email{ssanner@mie.utoronto.ca}, \email{elias.khalil@utoronto.ca}}

\maketitle              % typeset the header of the contribution
\begin{abstract}

Neural networks are being increasingly used as heuristics for constraint satisfaction. These neural methods are often recurrent, learning to iteratively refine candidate assignments. In this work, we make explicit the connection between such iterative neural heuristics and Large Neighborhood Search (LNS), and adapt an existing neural constraint satisfaction method---\methodname{}---into an LNS procedure. We decompose the resulting neural LNS into two standard components: the destroy and repair operators. On the destroy side, we instantiate several classical heuristics and introduce novel prediction-guided operators that exploit the model's internal scores to select neighborhoods. On the repair side, we utilize \methodname{} as a neural repair operator and compare the original sampling-based decoder to a greedy decoder that selects the most likely assignments. Through an empirical study on Sudoku, Graph Coloring, and MaxCut, we find that adapting the neural heuristic to an LNS procedure yields substantial gains over its vanilla settings and improves its competitiveness with classical and neural baselines. We further observe consistent design patterns across tasks: stochastic destroy operators outperform greedy ones, while greedy repair is more effective than sampling-based repair for finding a single high-quality feasible assignment. These findings highlight LNS as a useful lens and design framework for structuring and improving iterative neural approaches.

\keywords{Recurrent Transformer \and Large Neighborhood Search \and Constraint Satisfaction Problems.}
\end{abstract}

\section{Introduction}

Constraint satisfaction is ubiquitous in automated decision-making applications that involve planning, scheduling, and resource management, among others. By combining search, inference, and heuristics, constraint solvers have exhibited continual improvement over the past few decades, leading to reliable performance on a diverse range of challenging problems. That being said, there is also emerging interest in a complementary direction, namely the development of fast heuristics based on end-to-end machine learning (ML)~\cite{popescu2022overview,bengio2021machine}. 

By design, ML heuristics exploit statistical patterns within a homogeneous distribution of training instances, e.g., coloring problems on graphs with a similar structure. While an appropriately trained heuristic can perform well on test instances similar to those seen in training, it is unlikely to perform well out of distribution, e.g., when the number of nodes or the density of the graph changes in a coloring problem. One-shot solution prediction methods are particularly prone to this failure mode as they have  a fixed per-instance computational budget that cannot adapt to instance size or complexity~\cite{qiu2022dimes}. Classical solvers, on the other hand, simply search further, as desired.

To address this limitation and adapt to out-of-distribution instances, \textit{recurrence} has been introduced to many existing neural architectures. This includes masked diffusion models~\cite{shi2024simplified}, recurrent Transformers~\cite{fan2025looped,yang2023learning}, and single-step iterative Transformers~\cite{consformer}. In these approaches, the model is applied repeatedly over multiple iterations. The number of iterations can scale with the difficulty of the instance, partially bypassing the fixed-compute constraint of one-shot solution prediction and often improving generalization to harder or larger problems~\cite{consformer,fan2025looped}.
A common thread across these iterative neural methods is that they implicitly or explicitly refine a solution while optimizing some objective that captures the degree of constraint satisfaction. Viewed at a higher level, this is closely related to classical local search techniques and, in particular, Large Neighborhood Search (LNS)~\cite{lnsbook,shaw1998using}.

In this paper, we make this connection explicit by viewing a recent neural \textcolor{black}{constraint heuristic}, \methodname{}~\cite{consformer}, as an implicit neural LNS procedure. We formalize the deployment of \methodname{} within a generic LNS loop and decompose the resulting neural LNS into two standard components: the destroy operator and the repair operator.  \textcolor{black}{We refer to the resulting procedure as \methodname{}-LNS}. This perspective allows us to import standard LNS design dimensions and to ask: which classical LNS heuristics carry over to the neural setting, and what new heuristics become possible when the repair operator is a learned model with access to rich internal signals?
To this end, we make the following contributions:
\begin{itemize}
    \item We reinterpret the \methodname{}~\cite{consformer} as an implicit neural LNS solver in which the Transformer acts as part of the repair operator and embed it in an explicit LNS procedure.
    \item We propose prediction-guided destroy operators that exploit the neural network's logits as signals for variable selection, alongside classical heuristics.
    \item We conduct a systematic empirical study of classical and neural-specific LNS heuristics on several CSP benchmarks. Adapting the neural heuristic to an LNS procedure yields substantial gains over its vanilla iterative deployment.
\end{itemize}

\section{Related Work \& Background}

\subsection{\textcolor{black}{Constraint Satisfaction Problems}}

\textcolor{black}{
A \emph{constraint satisfaction problem} (CSP) is given by a tuple
$(\mathbf{X}, \mathcal{D}, \mathcal{C})$, where $\mathbf{X}=\{X_1,\dots,X_n\}$ is a set of variables, each $X_i$ takes values in a finite discrete domain $\mathcal{D}_i\in \mathcal{D}=\{\mathcal{D}_1,\dots,\mathcal{D}_n\}$, and $\mathcal{C}=\{c_1,\dots,c_m\}$ is a set of constraints. Each constraint $c_j \in \mathcal{C}$ is defined over a subset of variables $\mathbf{X}_j \subseteq \mathbf{X}$ and restricts the assignments to $\mathbf{X}_j$. Let $\mathcal{X} \triangleq \prod_{i=1}^n \mathcal{D}_i$ denote the Cartesian product over variable domains, and an assignment by $\mathbf{x}=(x_1,\dots,x_n) \in \mathcal{X}$. The objective of a CSP is to find $\mathbf{x}$ that satisfies every constraint:
$c_j(\mathbf{x}) = \texttt{true}, \quad \forall j \in [m]$.
}

\subsection{Large Neighborhood Search}
\label{sec:cspdef}
Large Neighborhood Search \textcolor{black}{is a popular approach for solving CSPs}~\cite{lnsbook,shaw1998using}. LNS iteratively destroys part of an existing solution and then rebuilds it. A destroy operator removes specific components of a solution. A repair operator then re-instantiates the removed elements to construct a new, complete feasible solution. 

Neural LNS has been an active area of recent research\textcolor{black}{~\cite{cappart2025combining,feng2025comprehensive}}. Existing work includes both utilizing neural methods as the destroy operator as well as the repair operator. For example, various works have studied learning a neural destroy operator to select the neighborhood in integer programs using reinforcement learning~\cite{wu2021learning,song2020general}, imitation learning~\cite{sonnerat2021learning}, contrastive learning~\cite{huang2023searching}, graph convolutional networks~\cite{zhou2023learning}, \textcolor{black}{and hindsight relabeling~\cite{feng2025spl}}. Hottung et al. use reinforcement learning to learn destroy and repair operators for vehicle routing problems~\cite{hottung2020neural,HOTTUNG2022103786,hottung2025neural}. Falkner et al.~\cite{falkner2022large} attack similar problems with graph neural networks as repair operators.

\subsection{Iterative Neural Heuristics}

Masked Generative Transformers~\cite{Chang_2022_CVPR}, Recurrent Transformers~\cite{dehghani2018universal}, and Masked Diffusion Models~\cite{shi2024simplified} have gained popularity for handling discrete data, and have often been applied to solve simple constraint satisfaction problems such as Sudoku~\cite{hamu2025accelerated,jolicoeur2025less,yang2023learning} and harder combinatorial optimization problems in general~\cite{sanokowski2024diffusion,sun2023difusco,consformer}.
Within these iterative neural heuristics, techniques like remasking and subset improvement have been used to enable iterative solution improvement where the framework forgets or refines parts of the solution~\cite{wang2025remasking,consformer}, which closely mirrors the LNS destroy-repair pattern.

\textcolor{black}{
We view an iterative neural heuristic as an update function $u$ that takes a complete but potentially infeasible assignment $\mathbf{x}$ together with a subset of variables $S \subseteq \mathbf{X}$ to be revised. We denote by
$
I := \{\, i \in \{1,\dots,n\} : X_i \in S \,\}
$
the corresponding set of indices and write
\[
\mathbf{x}' \sim u(\,\cdot \mid \mathbf{x}, I \,),
\qquad
x_i' = x_i \;\; \forall i \notin I.
\]
}

\begin{figure}
    \centering
    \includegraphics[width=0.8\linewidth]{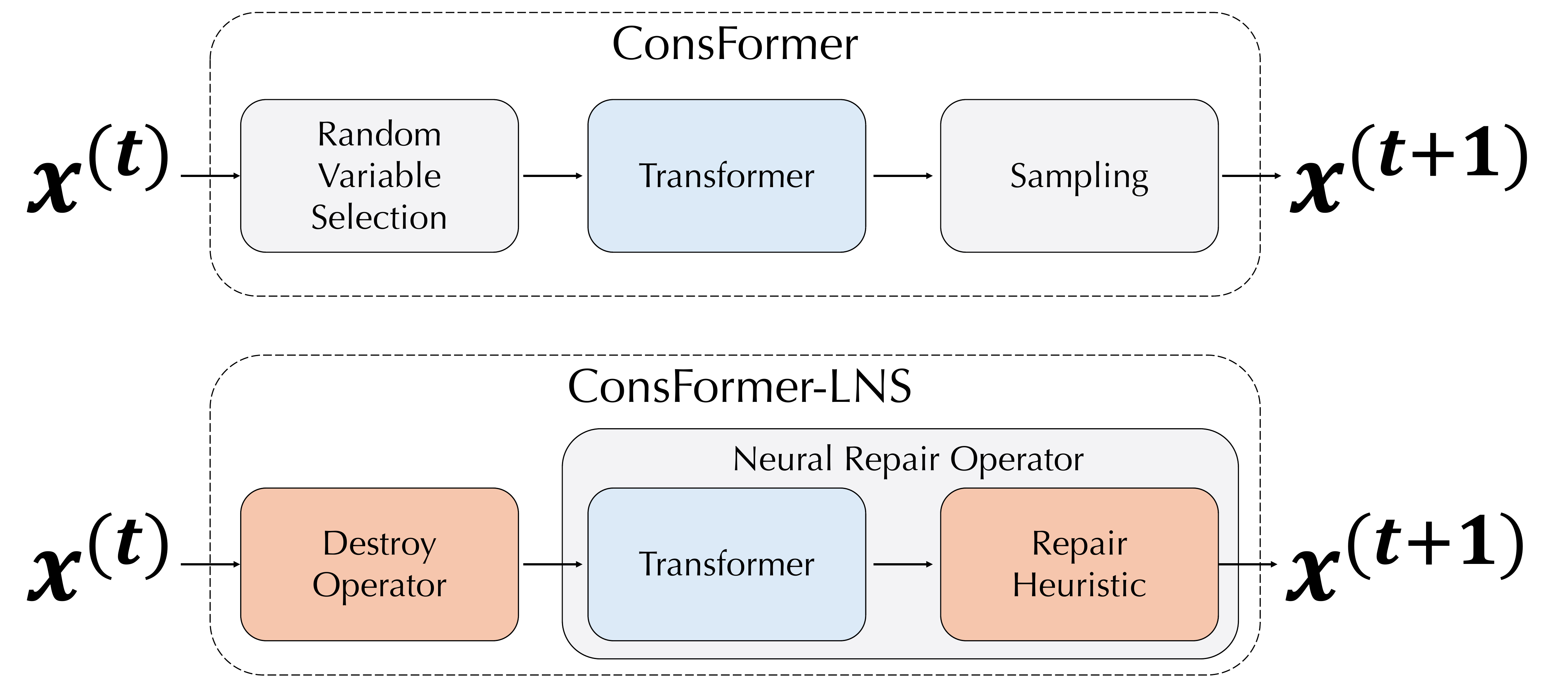}
    \caption{Comparison between the original \methodname{} update and our \methodname{}-LNS. \textbf{Top:} \methodname{} takes the current assignment $\mathbf{x}^{(t)}$, selects a random subset of variables, applies the Transformer, and samples a new assignment $\mathbf{x}^{(t+1)}$. \textbf{Bottom:} In \methodname{}-LNS, a destroy operator selects variables to modify, and the Transformer acts as part of a neural repair operator that proposes the next assignment $\mathbf{x}^{(t+1)}$.
    }

    \label{fig:framework}
\end{figure}

\subsubsection{ConsFormer.}

\label{sec:bg:consformer}

In this work, we adapt the~\methodname{}~\cite{consformer}, a recent model utilizing the Transformer architecture to solve constraint satisfaction problems.
\methodname{} \textcolor{black}{instantiates the update function $u$ by taking} in a complete, potentially infeasible variable assignment $\mathbf{x} = (x_1,\dots,x_n)$ as a sequence of tokens. Each variable's input representation combines a learned value embedding as well as positional information, i.e., information about the index $i$ of variable $X_i$.

The model randomly selects a subset of variables $S \subset \mathbf{X}$ to update during a forward pass. A specialized learnable embedding $\mathbf{e}_s$ is added to the tokens corresponding to variables in $S$ to signal their eligibility for modification. The Transformer processes this sequence to produce updated assignments; for variables in $S$, new values are sampled via the Gumbel-Softmax operator applied to the output logits to maintain differentiability, while variables not in $S$ remain unchanged.

\methodname{} is trained in a self-supervised way using differentiable constraint penalty functions $p_k$, such that $p_k(\mathbf{x})=0$ if $\mathbf{x}$ satisfies the $k$-th constraint and $p_k(\mathbf{x})>0$ otherwise. Given the model’s soft outputs for an instance, the training loss is a weighted sum of these penalties, $\mathcal{L}(\mathbf{x};\mathbf{\Theta}) = \sum_k \lambda_k f(p_k(\mathbf{x}))$, where $f$ is typically a quadratic transformation, weights $\lambda_k$ are hyperparameters, and $\mathbf{\Theta}$ are the learnable model parameters. This loss directly measures constraint violation under the model’s predictions and provides gradients with respect to the prediction scores.
In our work, we reuse \methodname{}’s architecture and self-supervised loss, and reinterpret it as part of a neural repair operator inside an LNS framework.

\section{Methodology}
\label{sec:method}

We reinterpret \methodname{} as a neural Large Neighborhood Search (LNS) procedure, allowing us to modify its components along the destroy and repair axes.

\subsection{Neural LNS View of ConsFormer}
\label{sec:lns_consformer}

We consider a CSP instance as defined in \cref{sec:cspdef}. A complete assignment is denoted by \(\mathbf{x} = (x_1,\dots,x_n)\), with a cost function \(\texttt{cost}(\mathbf{x})\) measuring constraint violation.
As described in \Cref{sec:bg:consformer}, \methodname{} iteratively refines an assignment \(\mathbf{x}^{(t)}\). We map this iterative process to the standard LNS components as follows:
\begin{itemize}
    \item \textbf{Destroy (The Subset Selection):} The random selection of the subset \(S^{(t)} \subseteq \mathbf{X} \) (and the application of the embedding \(\mathbf{e}_s\)) functions as a stochastic destroy operator. By flagging these variables for update, the model effectively ``unassigns'' them, rendering them eligible for modification while keeping \(\mathbf{X} \setminus S^{(t)}\) fixed. 

    \item \textbf{Repair (The Transformer Pass):} The Transformer's forward pass acts as a learned repair operator. Given the current assignment to the variables and the ``destroyed'' subset \(S^{(t)}\), the model proposes a new assignment $x_i^{(t+1)}$ for $X_i \in S^{(t)}$ by sampling from their corresponding output logits \(z_i^{(t)}\) via Gumbel-Softmax.
\end{itemize}

Figure~\ref{fig:framework} illustrates this reinterpretation: the original ConsFormer loop can be seen as an LNS procedure with a random destroy operator and a neural repair operator that samples for the next solution. We build on this view to study alternative design choices for each of the components.

\subsection{Destroy Operator}
\label{subsec:destroy}

At each iteration $t$, the destroy operator defines a binary mask $m^{(t)} \in \{0,1\}^n$,
where $m_i^{(t)} = 1$ means variable $X_i$ is selected. Following \methodname{}, the size of the selected set $S^{(t)} = \{ X_i : m_i^{(t)} = 1 \}$ adheres to a specified degree of destruction $\rho \in (0,1]$.
We adapt several classical destroy heuristics from the LNS literature~\cite{mara2022survey} and introduce neural \textcolor{black}{prediction-guided} methods that exploit internal signals from the ConsFormer model.

\subsubsection{Classical Destroy Operators}

\paragraph{Random removal.}

Let $\pi_i^{(t)} \in [0,1]$ denote the probability of destroying variable $X_i$ at iteration $t$.
The simplest strategy is to select the subset at random by setting all $\pi_i^{(t)}$ to the rate of destruction $\rho$, which corresponds to the original \methodname{}. We sample the binary mask by drawing a Bernoulli random variable for each variable:
$$
\pi_i^{(t)} = \rho, \quad
m_i^{(t)} \sim \mathrm{Bernoulli}(\pi_i^{(t)}).
$$

\paragraph{Greedy worst removal.}

A more targeted strategy destroys variables that contribute the most to constraint violations. Using \methodname{}'s relaxed constraint penalty loss $\mathcal{L}$, we define violation scores per variable to be $v_i(\mathbf{x}^{(t)})$ by evaluating the loss on the discretized variable assignments and extracting the per-variable contributions.
% computed in practice using the gradient of $p$ with respect to its one-hot encoding of $x_i^{(t)}$.
In practice, 
% let $\mathbf{u}^{(t)} = \text{OneHot}(\mathbf{x}^{(t)})$ and $\mathbf{u}_i^{(t)}$ be the one-hot vector for $x_i$; 
we compute $$v_i(\mathbf{x}^{(t)})=\left\|\frac{\partial \mathcal{L}\big(\text{OneHot}(\mathbf{x}^{(t)})\big)}{\partial \big(\text{OneHot}(\mathbf{x}^{(t)}_i)\big)}\right\|_1.$$
% $$
% \mathbf{v}_i^{(t)} \;=\; \frac{\partial L(\text{One-Hot}(\mathbf{x}^{(t)}))}
% {\partial \mathbf{x}_i^{(t)}}
% $$
\textcolor{black}{
We can then set $m_i^{(t)} = 1$ for the top-ranked variables with the largest scores (ties broken arbitrarily), yielding a greedy worst variable removal operator.
}

\paragraph{Stochastic worst removal.}
To improve diversification, we can randomize the selection while biasing towards highly violating variables~\cite{lnsbook}. We compute a normalized score from $v_i(\mathbf{x}^{(t)})$ such that the average probability matches $\rho$:
$$
\pi_i^{(t)} \propto v_i(\mathbf{x}^{(t)}), \quad \text{s.t.} \quad
\frac{1}{n} \sum_{i=1}^n \pi_i^{(t)} \approx \rho,
$$
$$m_i^{(t)} \sim \mathrm{Bernoulli}(\pi_i^{(t)})
$$
Variables with higher violation scores thus have a higher selection probability.

\paragraph{Stochastic related removal.}
Introduced by Shaw~\cite{shaw1998using}, related removal destroys a group of ``related'' variables at once. In our setting, a natural notion of relatedness is participation in a common constraint. At each iteration we
sample a random subset of constraints and then destroy all variables that belong to any selected constraint. Concretely, we draw a Bernoulli mask over constraints and define the destroy set accordingly:
$$
m_k^{\text{constr}} \sim \mathrm{Bernoulli}(\rho),\quad
S^{(t)} = \bigcup_{k : m_k^{\text{constr}} = 1} \mathrm{vars}(c_k).
$$
For graph-based problems where the number of constraints significantly exceeds the number of variables, we rescale the per-constraint Bernoulli parameter so that the expected number of selected constraints remains proportional to the desired variable-level degree of destruction $\rho$.

% $$
% c \sim \text{Unif}(\mathcal{C}), \quad S = \text{vars}(c),
% $$

\paragraph{Greedy related removal.}
We can build on random related removal by greedily selecting the constraints with the largest per constraint penalty $p_k$ introduced in \Cref{sec:bg:consformer}.
Given the current assignment $\mathbf{x}^{(t)}$, each constraint has a violation score $p_k(\mathbf{x}^{(t)}) \ge 0$, we select the subset of constraints with the largest penalties and destroy all variables that participate in them:
% $$
% \mathcal{K}^{(t)} = \text{top}_L \left(\bigl\{ p_k(\mathbf{x}^{(t)}) : c_k \in \mathcal{C} \bigr\}\right),
% \quad
% S^{(t)} = \bigcup_{k \in \mathcal{K}^{(t)}} \mathrm{vars}(c_k).
% $$

\textcolor{black}{
$$\mathcal{K}^{(t)}=\text{top}_L\!\left(\left\{p_k(\mathbf{x}^{(t)})\right\}_{c_k\in\mathcal{C}}\right),
\qquad
S^{(t)}=\bigcup_{k\in\mathcal{K}^{(t)}} \mathrm{vars}(c_k).
$$
where $\text{top}_L$ returns the indices of the constraints with the largest penalties, with ties broken arbitrarily. 
}

\subsubsection{Prediction-guided Destroy Operators}

We can design heuristics unique to the neural setting by utilizing the model's internal latent embeddings.
% The logits produced by ConsFormer's final hidden layer contain rich information about the model's beliefs over assignments.

\paragraph{Gradient-guided removal.}

After the final hidden layer, \methodname{} produces a vector of logits $z_i^{(t)} \in \mathbb{R}^{|\mathcal{D}_i|}$ corresponding to the variables $X_i$, which we can be interpreted as the model’s current belief over the values of $X_i$. Our first neural strategy uses the gradient of the loss function with respect to the model’s current belief to drive the destroy operator. 

To do this, we evaluate the penalty loss used during training on the current belief.  We then compute the gradient of $\mathcal{L}$ with respect to the logits
$$
g_i^{(t)} = \frac{\partial \mathcal{L}(\mathrm{softmax}(z^{(t)}))}{\partial z_i^{(t)}}
\quad\text{for each variable } X_i,
$$
and use the gradients as a per-variable score. We note that this is different from the gradient used for worst removal, since we rely on the model's internal signals instead of a discretized variable assignment. We include a greedy and a stochastic variant similar to classical approaches, where the greedy variant selects the top-ranked variables with the largest gradients, and the random variant defines 
$$
\pi_i^{(t)} \propto \big\lVert g_i^{(t)} \big\rVert_1, \quad \text{s.t.} \quad
\frac{1}{n} \sum_{i=1}^n \pi_i^{(t)} \approx \rho,
$$
$$m_i^{(t)} \sim \mathrm{Bernoulli}(\pi_i^{(t)}) \, .
$$ 

Variables with larger $\pi_i^{(t)}$ are those for which small changes in the prediction would most strongly affect the loss. In contrast to purely violation-based heuristics, this gradient-based removal explicitly leverages the neural model and the training loss to identify variables that are most ``responsible'' for the current soft constraint violations. 
% Intuitively, an assignment that is both highly violated and predicted with high confidence by the model is a particularly problematic mistake.

\paragraph{Confidence-margin removal.}

Inspired by recent work in Masked Diffusion Models~\cite{kim2025train,hamu2025accelerated}, we introduce a strategy using solely the model's current belief. Intuitively, we target variables for which the model is highly uncertain, regardless of their current violation.

We first transform the logits into probabilities 
% $p_i(v) = \mathrm{softmax}\big(z_i^{(t)}\big)_v, \quad v \in \mathcal{D}_i,$ 
$q_i^{(t)} = \mathrm{softmax}\big(z_i^{(t)}\big),$ 
we then quantify confidence using the gap between the top two probabilities in $q_i^{(t)}$. Let $v_1,v_2$
% $q_i^{(1)} \ge q_i^{(2)}$ 
be the assignment values with the largest and second-largest entries in $q_i$, we define a confidence margin
$$
\mathrm{margin}_i^{(t)} = q_i^{(t)}(v_1) - q_i^{(t)}(v_2).
$$
This margin can be used as the score for selecting the variables. Intuitively, small margins indicate that the model is unsure between multiple values. We again introduce a greedy and a stochastic variant where $\pi_i^{(t)} \propto (-margin_i^{(t)}).$This encourages the search to repeatedly revisit parts of the assignment where the model has low confidence and may benefit from additional refinement.

\subsection{Repair Operator}
\label{subsec:repair}

The repair operator takes the current assignment \(\mathbf{x}^{(t)}\) and destroy set \(S^{(t)}\), and proposes a candidate assignment \(\mathbf{x}^{(t+1)}\) by modifying only variables in \(S^{(t)}\). In our framework, \methodname{} is utilized as a neural repair operator. Given \(\mathbf{x}^{(t)}\) and mask \(m^{(t)}\), the model produces logits and corresponding probabilities $q_i^{(t)}(v)$ for each variable $x_i^{(t)}$ and value assignment $v$. We consider two decoding strategies: sampling and greedy decoding.

% todo: put consformer architecture here?

\paragraph{Stochastic sampling.}  
This is the original ConsFormer implementation. For each variable with \(m_i^{(t)} = 1\), we sample $x_i^{(t+1)} \sim q_i^{(t)}(v)$ using the Gumbel--Softmax sampler; for \(m_i^{(t)} = 0\), we keep $x_i^{(t+1)} = x_i^{(t)}$. This yields a stochastic repair operator that explores the neighborhood induced by the destroy mask.

\paragraph{Greedy decoding.}  
A deterministic alternative selects the most likely value for each variable with \(m_i^{(t)} = 1\):
$$
x_i^{(t+1)} \in \arg\max_{v \in \mathcal{D}_i} q_i^{(t)}(v)
$$
This corresponds to a greedy repair step based on the model’s current beliefs. 

\section{Experiments}

This section aims to empirically answer the following
research questions:

\begin{itemize}

    \item \textbf{RQ1 (ConsFormer vs.\ ConsFormer-LNS):}
    Does adapting ConsFormer to a neural LNS procedure improve its performance? 
    
    \item \textbf{RQ2 (Classical vs.\ Prediction-guided Destroy Operators):}
    How do classical and prediction-guided destroy operators contribute to performance gains?

    \item \textbf{RQ3 (Greedy vs.\ Stochastic Destroy Operators):} How important is randomness in the destroy operator?

    \item \textbf{RQ4 (Greedy vs.\ Stochastic Repair Operators):}
    Given a fixed destroy strategy, does greedy repair generally outperform the original sampling-based repair, and how does its per-instance behavior differ from sampling?

    \item \textbf{RQ5 (\methodname{}-LNS vs. Other Solvers):}
    How does the LNS-enhanced ConsFormer perform compared to other neural or classical solvers?
\end{itemize}

\subsection{Experimental Setup}

\subsubsection{Problem Selection}

We evaluate on the same datasets as Xu et al.~\cite{consformer}, excluding the nurse rostering problem for which \methodname{} already solves 100\% of the instances.

\paragraph{Sudoku} is a simple CSP that involves filling a $9 \times 9$ grid with digits from 1 to 9 such that each row, column, and $3 \times 3$ sub-grid contains all 9 numbers. A single Sudoku instance is defined by a partially filled board and its difficulty is determined by the number of initial values: fewer initially filled cells in the board involve a larger space of possible assignments to the unfilled cells and is therefore harder. Following Xu et al.~\cite{consformer}, we use the dataset from SATNet~\cite{satnet} for training and in-distribution testing, and the dataset from RRN~\cite{rrn} for harder out-of-distribution testing. 

\paragraph{Graph Coloring} seeks an assignment of colors to vertices in a graph such that no two neighboring nodes share the same color. The problem is defined by the graph's structure and the number of available colors $k$. We study the same 5-coloring and 10-coloring datasets, where training graphs have 50 vertices for $k=5$ and 100 vertices for $k=10$ whereas OOD graphs have 100 for $k=5$ and 200 for $k=10$.

\paragraph{MaxCut} aims to identify a cut in a graph such that the number of edges crossing the partition is maximized. \textcolor{black}{MaxCut can be alternatively viewed as a 2-coloring Max-CSP which seeks an assignment that maximizes the number of satisfied constraints. Since our framework works by reducing the number of violated constraints, it can be applied directly for MaxCut.}
The model is trained on small generated graphs and tested on benchmark instances from the GSET dataset~\cite{ye2003gset}, which includes graphs with sizes ranging from 800 to 10000 vertices.

\subsubsection{Baselines}
For RQ5, we compare against a robust set of neural and classical baselines established in the literature following Xu et al.~\cite{consformer}. OR-Tools~\cite{cpsat}\footnote{\textcolor{black}{All OR-Tools results are reported using version 9.10 and were run on the same hardware as the \methodname{}-LNS models.}}, a state-of-the-art constraint programming solver, is the primary non-learning baseline; note that many of its internal sub-solving routines utilize Large Neighborhood Search (LNS)~\cite{cpsat}. We omit OR-Tools from the Sudoku experiments, as these instances are computationally trivial for exact solvers and are primarily used to benchmark the reasoning capabilities of neural approaches.

\subsubsection{Implementation Details}

We train our models on single H100 GPU nodes. We adopt the hyperparameter configurations from Xu et al.~\cite{consformer} and \textcolor{black}{retrain a separate model for each destroy/repair operator pairing}\footnote{\textcolor{black}{Our code is available at \url{https://github.com/khalil-research/ConsFormer}.}}.

\subsection{Results \& Analysis}

Experimental results comparing the different destroy and repair operators are presented in \Cref{tab:sudoku_transposed,tab:maxcut_transposed_v3,tab:coloring_combined_transposed}. Italic numbers correspond to the original ConsFormer configuration (random destroy + sampling-based repair); bold and underlined numbers mark the best and second best result within each dataset, respectively.

\begin{table}
\centering
\caption{
Sudoku instances solved (\%) for different combinations of destroy and repair operators. Columns correspond to the destroy operators specified in \Cref{subsec:destroy} where the Greedy and Stochastic variants are labeled (Gr.) and (St.) respectively. Rows vary the repair operator (sampling-based vs. greedy). Test instances contain $1,000$ instances from the SATNet dataset, OOD refers to Out-of-Distribution evaluation on the RRN test dataset which contains 18K instances. All configurations are evaluated for 2K iterations.}

\label{tab:sudoku_transposed}
\setlength{\aboverulesep}{0pt}
\setlength{\belowrulesep}{0pt}
\renewcommand{\arraystretch}{1.1} % Adds the breathing room back

% \resizebox{\textwidth}{!}{%
\begin{tabular}{ll@{\hskip 0.1in}|@{\hskip 0.1in}ccccccccc}
\toprule
\multirow{2}{*}{\textbf{Dataset}} & \multirow{2}{*}{\textbf{Repair}} & \multirow{2}{*}{\textbf{Random}} & \multicolumn{2}{c}{\textbf{Worst}} & \multicolumn{2}{c}{\textbf{Related}} & \multicolumn{2}{c}{\textbf{Gradient}} & \multicolumn{2}{c}{\textbf{Confidence}} \\
\cmidrule(lr){4-5} \cmidrule(lr){6-7} \cmidrule(lr){8-9} \cmidrule(lr){10-11}
& & & \textbf{Gr.} & \textbf{St.} & \textbf{Gr.} & \textbf{St.} & \textbf{Gr.} & \textbf{St.} & \textbf{Gr.} & \textbf{St.} \\
\midrule
\multirow{2}{*}{\textbf{Test}} & Sampled & \textbf{\textit{100}} & 94.3 & \textbf{100} & 99.7 & \textbf{100} & 20.6 & \textbf{100} & 0.0 & \textbf{100} \\
& Greedy & \textbf{100} & 85.6 & \textbf{100} & \underline{99.9} & \textbf{100} & 0.0 & \textbf{100} & 0.0 & \textbf{100}\\
\midrule
\multirow{2}{*}{\textbf{OOD}} & Sampled & \textit{85.8} & 16.7 & 71.7 & 32.1 & 48.7 & 0.1 & 90.6 & 0.0 & 90.8 \\
& Greedy & 81.2 & 11.8 & \textbf{91.8} & 30.9 & 63.9 & 0.1 & 84.1 & 0.0 & \underline{91.5} \\
\bottomrule
\end{tabular}%
% }
\end{table}

\begin{table}
\centering
\caption{MaxCut performance on GSET: average gap to the best known cut (lower is better) for different graph sizes. Columns and Rows correspond to destroy and repair operators as in \Cref{tab:sudoku_transposed}. All configurations are evaluated with a 180s time limit.}
\label{tab:maxcut_transposed_v3}

\setlength{\aboverulesep}{0pt}
\setlength{\belowrulesep}{0pt}
\renewcommand{\arraystretch}{1.1}

\resizebox{\textwidth}{!}{%
\begin{tabular}{ll@{\hskip 0.1in}|@{\hskip 0.1in}ccccccccc}
\toprule
\multirow{2}{*}{\textbf{Size}} & \multirow{2}{*}{\textbf{Repair}} & \multirow{2}{*}{\textbf{Random}} & \multicolumn{2}{c}{\textbf{Worst}} & \multicolumn{2}{c}{\textbf{Related}} & \multicolumn{2}{c}{\textbf{Gradient}} & \multicolumn{2}{c}{\textbf{Confidence}} \\
\cmidrule(lr){4-5} \cmidrule(lr){6-7} \cmidrule(lr){8-9} \cmidrule(lr){10-11}
 & & & \textbf{Gr.} & \textbf{St.} & \textbf{Gr.} & \textbf{St.} & \textbf{Gr.} & \textbf{St.} & \textbf{Gr.} & \textbf{St.} \\
\midrule
\multirow{2}{*}{\textbf{$|V|{=}800$}} & Sampled & \textit{16.33} & 430.11 & 473.00 & 923.67 & 472.33 & 167.78 & 45.11 & 489.67 & \underline{9.78} \\
 & Greedy & 31.67 & 503.11 & 38.56 & 915.00 & 71.33 & 137.00 & \textbf{4.44} & 397.00 & 21.00 \\
\midrule
\multirow{2}{*}{\textbf{$|V|{=}1K$}} & Sampled & \textit{12.44} & 382.89 & 417.67 & 781.00 & 418.56 & 154.89 & 30.89 & 469.22 & \underline{9.78} \\
 & Greedy & 25.00 & 414.11 & 33.22 & 714.33 & 57.00 & 116.56 & \textbf{8.00} & 364.11 & 18.78 \\
\midrule
\multirow{2}{*}{\textbf{$|V|{=}2K$}} & Sampled & \textit{52.11} & 832.89 & 944.78 & 1646.11 & 944.67 & 297.00 & 69.89 & 1014.00 & \underline{37.11} \\
 & Greedy & 74.67 & 842.89 & 97.00 & 1472.33 & 157.33 & 207.44 & \textbf{30.56} & 674.00 & 62.78 \\
\midrule
\multirow{2}{*}{\textbf{$|V|\geq3K$}} & Sampled & \textit{115.25} & 1268.75 & 1618.00 & 2311.50 & 1467.38 & 530.63 & 122.75 & 1921.13 & \underline{106.63} \\
 & Greedy & 142.50 & 1286.50 & 188.00 & 2085.50 & 229.88 & 354.50 & \textbf{63.63} & 1230.13 & 140.25 \\
\bottomrule
\end{tabular}%

}
\end{table}

\begin{table}
\centering
\caption{Graph-Coloring instances solved (\%). Top: Coloring-5, Bottom: Coloring-10. Columns and Rows correspond to destroy and repair operators as in \Cref{tab:sudoku_transposed}. OOD refers to Out-of-Distribution evaluation for ANYCSP and \methodname{} where the number of vertices $n$ in the graph is larger than that of the training instances. All datasets have 1200 instances. All configurations are evaluated with a 10s time limit.}
\label{tab:coloring_combined_transposed}

\setlength{\aboverulesep}{0pt}
\setlength{\belowrulesep}{0pt}
\renewcommand{\arraystretch}{1.1}

\begin{tabular}{ll@{\hskip 0.1in}|@{\hskip 0.1in}ccccccccc}
\toprule
\multirow{2}{*}{\textbf{Dataset}} & \multirow{2}{*}{\textbf{Repair}} & \multirow{2}{*}{\textbf{Random}} & \multicolumn{2}{c}{\textbf{Worst}} & \multicolumn{2}{c}{\textbf{Related}} & \multicolumn{2}{c}{\textbf{Gradient}} & \multicolumn{2}{c}{\textbf{Confidence}} \\
\cmidrule(lr){4-5} \cmidrule(lr){6-7} \cmidrule(lr){8-9} \cmidrule(lr){10-11}
& & & \textbf{Gr.} & \textbf{St.} & \textbf{Gr.} & \textbf{St.} & \textbf{Gr.} & \textbf{St.} & \textbf{Gr.} & \textbf{St.} \\
\midrule
\multicolumn{11}{c}{\textbf{Graph-Coloring-5} ($n=50 \rightarrow n=100$)} \\
\midrule
\multirow{2}{*}{\textbf{Test}} & Sampled & \textit{81.6} & 33.4 & 77.8 & 32.0 & 81.8 & 41.7 & 79.9 & 25.7 & 82.3 \\
& Greedy & \textbf{82.9} & 32.9 & 82.5 & 22.1 & \underline{82.8} & 41.2 & \textbf{82.9} & 32.7 & 81.8 \\
\midrule
\multirow{2}{*}{\textbf{OOD}} & Sampled & \textit{46.3} & 0.0 & 41.0 & 0.0 & 47.6 & 0.7 & 44.8 & 0.0 & 49.0 \\
& Greedy & \textbf{54.2} & 0.1 & 52.5 & 0.0 & 53.7 & 2.6 & \underline{54.0} & 0.0 & 49.4 \\
\midrule
\multicolumn{11}{c}{\textbf{Graph-Coloring-10} ($n=100 \rightarrow n=200$)} \\
\midrule
\multirow{2}{*}{\textbf{Test}} & Sampled & \textit{53.6} & 0.0 & \underline{53.8} & 0.0 & 53.5 & 11.9 & \textbf{53.9} & 0.0 & \underline{53.8} \\
& Greedy & \textbf{53.9} & 0.4 & \textbf{53.9} & 0.0 & \textbf{53.9} & 14.8 & \textbf{53.9} & 5.5 & \textbf{53.9} \\
\midrule
\multirow{2}{*}{\textbf{OOD}} & Sampled & \textit{10.2} & 0.0 & 14.7 & 0.0 & 14.2 & 2.6 & 14.5 & 0.0 & 14.8 \\
& Greedy & \underline{17.4} & 0.0 & 16.8 & 0.0 & \textbf{18.4} & 5.1 & 16.6 & 1.0 & 16.3 \\
\bottomrule
\end{tabular}%
% }
\end{table}

\begin{figure}
    \centering
    \includegraphics[width=0.71\linewidth]{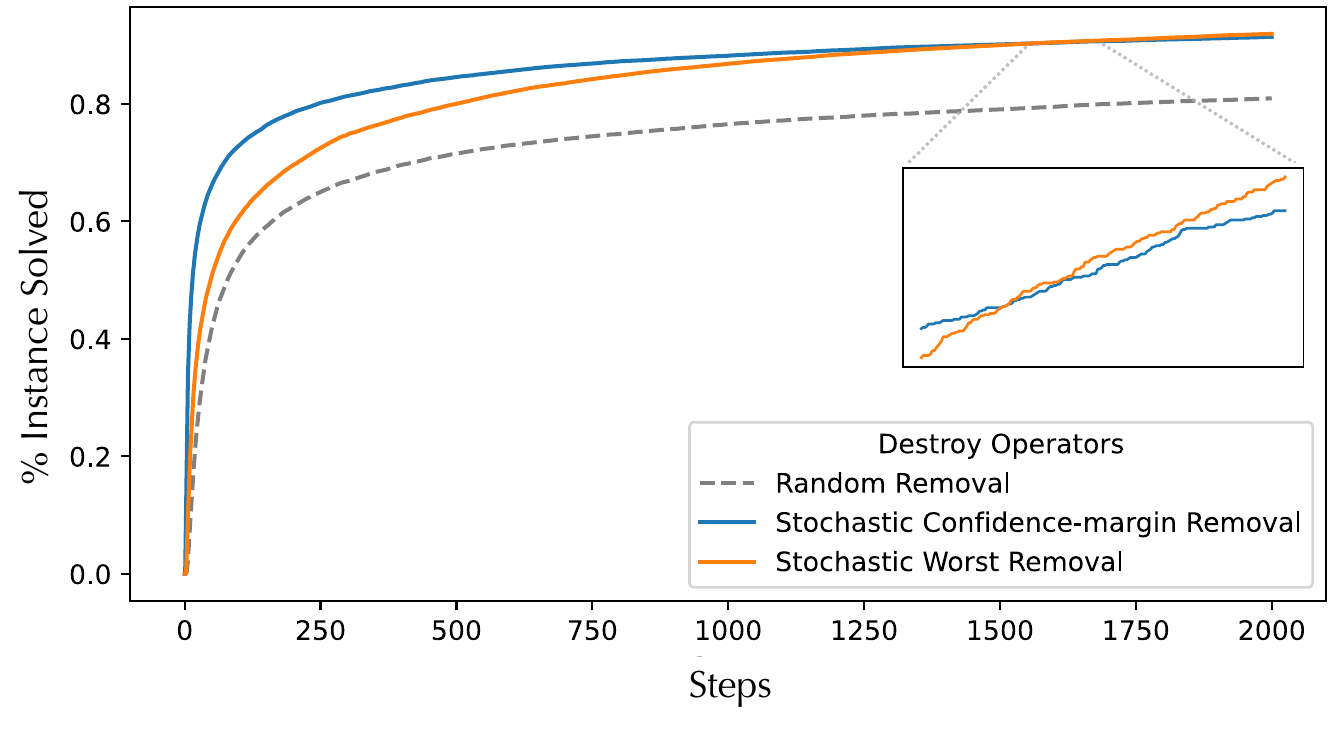}
    % \caption{Comparison of the best performing Classical and Prediction-guided Destroy Operators.}
    \caption{Sudoku instances solved (\%) over LNS steps for the baseline random destroy and the best-performing classical (stochastic worst removal) and prediction-guided (stochastic confidence-margin removal) destroy operators. Both heuristics outperform random removal, with confidence-margin removal improving fastest early on and stochastic worst removal eventually achieving the highest solved percentage.}

    \label{fig:sudokudestroys}
\end{figure}

\subsubsection{RQ1: ConsFormer vs.\ ConsFormer-LNS.}

Across all benchmarks, adapting \methodname{} into an explicit LNS procedure enhances model performance. On Sudoku (\Cref{tab:sudoku_transposed}), \methodname{}-LNS improves the out-of-distribution (OOD) percentage of instances solved from $85.8\%$ to $91.8\%$. On MaxCut (\Cref{tab:maxcut_transposed_v3}), \methodname{}-LNS reduces the gap to the best known cut sizes from $16.33$ to $4.44$ for $|V|=800$, from $12.44$ to $8.00$ for $|V|=1K$, from $52.11$ to $30.56$ for $|V|=2K$, and from $115.25$ to $63.63$ for $|V|\geq3K$. On graph coloring (\Cref{tab:coloring_combined_transposed}), OOD percentage of instances solved increased from $46.3\%$ to $54.2\%$ for $k=5$ and from $10.2\%$ to $18.4\%$ for $k=10$.

\subsubsection{RQ2: Classical vs.\ Prediction-guided destroy.}

Overall, the best performing classical and prediction-guided destroy operators achieve similar performance gain from the baseline, with the exception of MaxCut, where the gradient-guided removal achieves much stronger performance. We further compare the best performing methods by examining the instance solved percentage across iterations in~\Cref{fig:sudokudestroys}. We observe that the confidence-based removal rapidly improves in early iterations, but is overtaken by the random worst removal operator towards the end. 

The strong early performance of the confidence-margin model is consistent with the intuition that the model is prioritizing the variable assignments it has the least confidence in. However, this destroy operator receives no explicit signal from the current solution's quality, therefore, as the model becomes more certain in the later iterations, its improvement slows down and eventually gets overtaken by methods with richer penalty-derived signals. This observation aligns with the efficiency-accuracy trade-off reported in the Masked Diffusion literature~\cite{hamu2025accelerated}. 

The lack of a clear winner among destroy operators, across all benchmarks, is in line with the findings of classical LNS literature:  problem structure determines the most suitable Destroy operator~\cite{lnsbook}. Further augmentation of \methodname{}-LNS with an adaptive mechanism~\cite{mara2022survey} is therefore a promising direction for future work.
%be beneficial which we leave for future work. 

% \todo{draw a figure of the pleateauing accuracy for the different variants}
\begin{figure}[t]
    \centering
    \includegraphics[width=\linewidth]{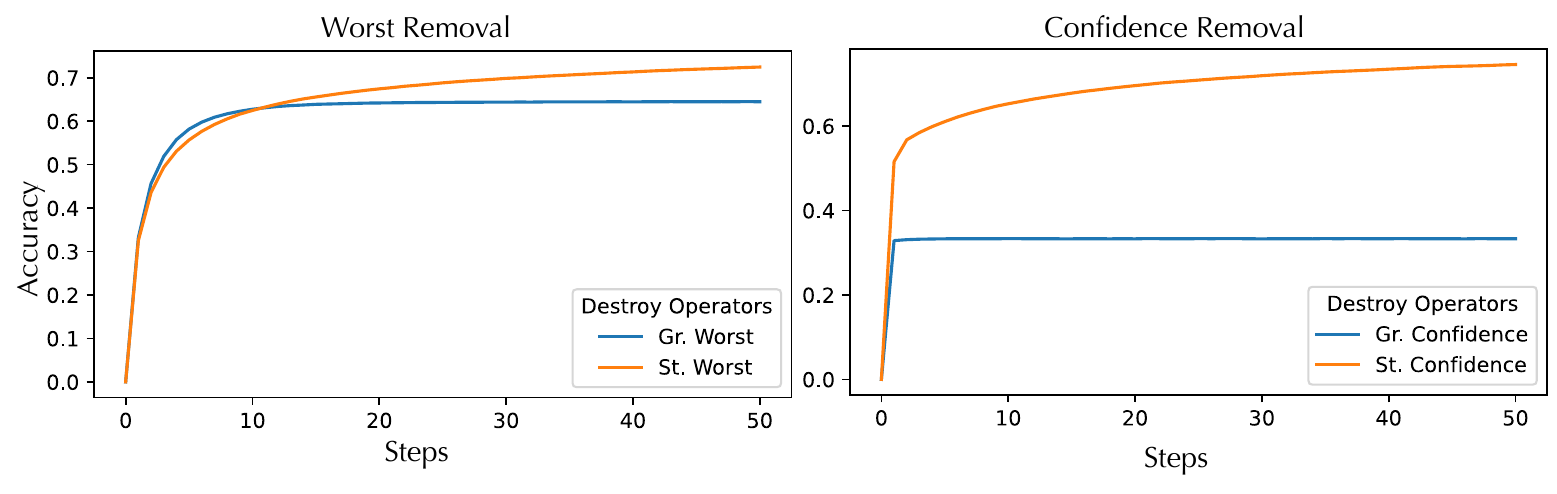}
    % \caption{Comparison of greedy vs stochastic variants of the different destroy operators.}
    \caption{Cell accuracy for Sudoku over iterations for greedy (blue) and stochastic (orange) variants of Worst and Confidence destroy operators. Greedy variants plateau early, while stochastic variants continue to improve and reach substantially higher accuracies.}
    \label{fig:greedyvstochastic}
\end{figure}

\subsubsection{RQ3: Greedy vs.\ Stochastic Destroy Operators}

The failure of purely greedy destroy operators is observed across all benchmarks.  Greedy variants of the different strategies often collapse to near-zero solved percentages on OOD Sudoku and coloring and exhibit very poor cut values on MaxCut. This echoes the classical LNS literature, where randomization is often needed to escape local optima. In contrast, the randomized variants tend to perform well, highlighting the importance of controlled randomness in the destroy step. \Cref{fig:greedyvstochastic} shows the per-cell accuracy scores of the best performing destroy operators, i.e., the fraction of variables that are assigned to the correct value in the groundtruth unique feasible Sudoku solution. It can be seen that the greedy variant plateaus much earlier while the stochastic variants are able to achieve much higher scores.

\subsubsection{RQ4: Greedy vs.\ Stochastic Repair Operators}
\begin{figure}[!h]
    \centering
\includegraphics[width=0.92\linewidth]{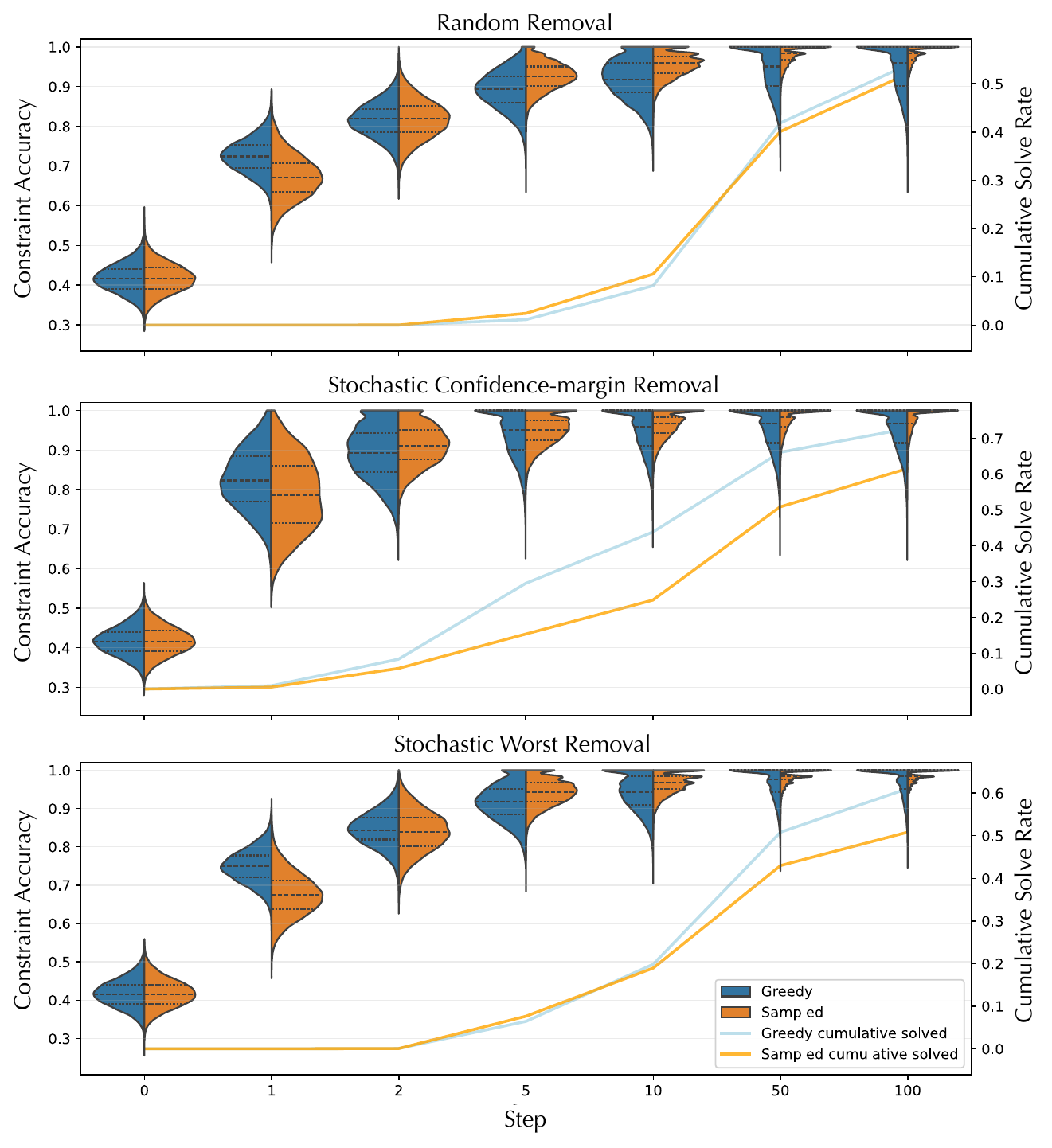}
    % \caption{Sudoku Constraint Accuracy Sampling vs Greedy Repair Operator}
    \caption{Sudoku constraint accuracy on OOD Test dataset over LNS steps for greedy (blue) and sampling-based (orange) repair under Random and two best-performing destroy operators (Stochastic Confidence-margin, Stochastic Worst). Split violin plots show the distribution of per-instance constraint satisfaction at selected steps, while the lines on the right axis show the cumulative fraction of instances solved. Greedy repair achieves higher final solved percentages but exhibits a wider tail of unsolved instances, whereas sampling attains higher average constraint accuracy at intermediate steps.
}

    \label{fig:sudokurepairs}
\end{figure}

The greedy repair operator almost always outperforms stochastic sampling-based repair across all benchmarks. To better understand this result, we run \methodname{}-LNS with the baseline and best-performing destroy operators for a small number of iterations for Sudoku. We plot the distribution of the constraint-satisfaction rate as well as the cumulative percentage of instance solved in \Cref{fig:sudokurepairs}. Across all three configurations, we observe a few clear patterns. 

The Greedy repair operator has a better improvement in the first step, but gets overtaken by the stochastic repair operator quickly. At later iterations, the stochastic repair operator has a better overall distribution with a mean closer to 100\% accuracy, while greedy repair has a long tail of unsolved instances and a spike at 100\% accuracy. Despite the worse performance per step, the greedy repair operator consistently achieves a higher cumulative instance solved percentage in the later steps.

This suggests that while the stochastic sampling-based repair operator learns the solution distribution better through exploration, the greedy repair operator can find the single best solution more often. Intuitively, the superior performance of the greedy operator in our setting is due to our objective of finding a single best solution minimizing our constraint penalty. Our evaluation metrics for the tasks only depend on the best solution and not on the quality or diversity of the underlying solution distribution. Given that the neural model is trained to minimize a constraint violation penalty for the assignments, greedy repair is effectively a \textit{maximum a posteriori}---or MAP---decoding step that directly targets a single high-quality assignment, whereas sampling is better aligned with exploring multiple diverse assignments.

\begin{table}[]
\centering
\caption{Performance comparison for Sudoku. In-distribution test instances contain $1,000$ instances; OOD refers to RRN test dataset (18K instances). For all but \methodname{} and \methodname{}-LNS, the reported instance solved (\%) are based on the results from Xu et al.~\cite{consformer}.}
\label{tab:sudoku_vs_others_transposed}

\setlength{\aboverulesep}{0pt}
\setlength{\belowrulesep}{0pt}
\renewcommand{\arraystretch}{1.1}

\resizebox{0.8\textwidth}{!}{%
\begin{tabular}{l@{\hskip 0.1in}|@{\hskip 0.1in}cccccc}
\toprule
\textbf{Dataset} & \textbf{SATNet} & \textbf{RRN} & \textbf{Recurrent} & \textbf{IRED} & \textbf{\methodname{}} & \textbf{\methodname{}} \\
 & \cite{satnet} & \cite{rrn} & \textbf{Trans.}\cite{yang2023learning} & \cite{ired} & & \textbf{-LNS} \\
\midrule
Test & 98.3 & \underline{99.8} & \textbf{100} & 99.4 & \textbf{100} & \textbf{100} \\
OOD & 3.2 & 28.6 & 32.9 & 62.1 & \underline{85.8} & \textbf{91.8} \\
\bottomrule
\end{tabular}%
}
\end{table}

\begin{table}
\centering
\caption{Performance comparison for Graph-Coloring tasks. All methods are evaluated with a 10s time limit. OOD refers to Out-of-Distribution evaluation where $n$ is larger than training instances. For all but \methodname{} and \methodname{}-LNS, the reported instance solved (\%) are based on the results from Xu et al.~\cite{consformer}.}
\setlength{\aboverulesep}{0pt}
\setlength{\belowrulesep}{0pt}
\renewcommand{\arraystretch}{1.1}

\resizebox{0.8\textwidth}{!}{
\begin{tabular}{l@{\hskip 0.1in}|@{\hskip 0.1in}ccccc}
\toprule
\textbf{Dataset} & \textbf{Greedy} & \textbf{OR-Tools} & \textbf{ANYCSP} & \textbf{\methodname{}} & \textbf{\methodname{}} \\
& & \cite{cpsat} & \cite{anycsp} & & \textbf{-LNS} \\
\midrule
\multicolumn{6}{c}{\textbf{Graph-Coloring-5} ($n=50 \rightarrow n=100$)} \\
\midrule
Test & 32.42 & \textbf{83.08} & 79.17 & 81.60 & \underline{82.90} \\
OOD & 0.00 & \textbf{57.16} & 34.83 & 47.33 & \underline{54.20} \\
\midrule
\multicolumn{6}{c}{\textbf{Graph-Coloring-10} ($n=100 \rightarrow n=200$)} \\
\midrule
Test & 0.75 & 52.41 & 0.00 & \underline{53.60} & \textbf{53.90} \\
OOD & 0.00 & \underline{10.25} & 0.00 & 10.20 & \textbf{18.40} \\
\bottomrule
\end{tabular}
}
\label{table:coloring_vs_others_transposed}
\end{table}

\begin{table}
\centering
\caption{Performance comparison for MaxCut tasks on GSET. Numbers reported are the average gap to the best known cut size, the lower the better. Values are as reported in Xu et al.\cite{consformer}. We similarly set a time limit of 180 seconds.}
\label{tab:maxcut_vs_others_transposed}

\setlength{\aboverulesep}{0pt}
\setlength{\belowrulesep}{0pt}
\renewcommand{\arraystretch}{1.1}

\resizebox{\textwidth}{!}{%
\begin{tabular}{l@{\hskip 0.1in}|@{\hskip 0.1in}ccccccccc}
\toprule
\textbf{Size} & \textbf{Greedy} & \textbf{SDP} & \textbf{RUNCSP} & \textbf{ECO-DQN} & \textbf{ECORD} & \textbf{ANYCSP} & \textbf{OR-Tools} & \textbf{\methodname{}} & \textbf{\methodname{}} \\
 & & \cite{sdp} & \cite{runcsp} & \cite{ECODQN} & \cite{ecord}& \cite{anycsp} & \cite{cpsat} & & \textbf{-LNS} \\
\midrule
$|V|{=}800$ & 411.44 & 245.44 & 185.89 & 65.11 & 8.67 & \textbf{1.22} & 143.89 & 16.33 & \underline{4.44} \\
$|V|{=}1K$ & 359.11 & 229.22 & 156.56 & 54.67 & 8.78 & \textbf{2.44} & 112.78 & 12.44 & \underline{8.0} \\
$|V|{=}2K$ & 737.00 & - & 357.33 & 157.00 & 39.22 & \textbf{13.11} & 365.89 & 52.11 & \underline{30.56} \\
$|V|{\geq}3K$ & 774.25 & - & 401.00 & 428.25 & 187.75 & \textbf{51.63} & 378.62 & 115.25 & \underline{63.63} \\
\bottomrule
\end{tabular}%
}
\end{table}

\subsubsection{RQ5 \methodname{}-LNS vs. Other Solvers}

We compare the best \methodname{}-LNS configuration on each benchmark against classical and neural baselines (\Cref{tab:sudoku_vs_others_transposed,table:coloring_vs_others_transposed,tab:maxcut_vs_others_transposed}). Overall, enhancing ConsFormer with LNS substantially improves its position among existing methods. On Sudoku, \methodname{}-LNS is compared to other neural approaches and attains the highest OOD percentage of instance solved ($91.8\%$ vs.\ $62.1\%$ for IRED~\cite{ired}), while matching the best methods on the in-distribution test set. For graph coloring with $k=5$, \methodname{}-LNS improves on \methodname{} but is still slightly behind OR-Tools on the smaller instances. On the harder graph coloring tasks with $k=10$, it becomes strongest overall. It marginally surpasses OR-Tools on the test set and achieves a substantially higher OOD solved percentage ($18.4\%$ vs.$10.25\%$). For MaxCut, \methodname{}-LNS consistently improves upon ConsFormer, achieving second-best performance across all graph sizes.

\section{Conclusion}

In this work, we have made the connection between iterative neural constraint solvers and Large Neighborhood Search (LNS) explicit. We reinterpreted \methodname{}, a recent neural heuristic, as a neural repair operator and adapted it into an LNS procedure. We implemented classical as well as novel prediction-guided destroy operators that leverage the model's internal confidence and gradient signals. We systematically evaluated combinations of destroy and repair strategies.

Our empirical evaluation on Sudoku, Graph Coloring, and MaxCut demonstrated a few key findings:
\begin{itemize}
    \item Porting the \methodname{} into the LNS framework yielded substantial performance gains across all benchmarks.
    \item Echoing classical LNS findings, purely greedy destroy operators frequently collapsed to poor local minima, whereas their stochastic counterparts continued to improve over iterations and reached much better final performance.
    \item Our novel destroy operators showed strong performance in early iterations and specific tasks like MaxCut, but no single destroy operator dominated across all tasks. This suggests that an Adaptive LNS~\cite{mara2022survey,ropke2006adaptive} that utilizes multiple destroy operators is a promising direction for future work.
    \item Although sampling-based repair attained better average constraint satisfaction at intermediate steps, greedy decoding almost always achieved higher overall performance, consistent with a MAP-style decoding optimizing for a single best assignment rather than exploring a diverse solution distribution.
    \item Compared to other approaches, \methodname{}-LNS attains the strongest OOD performance among neural Sudoku solvers, becomes competitive with OR-Tools on graph coloring and strongest on the hardest $k{=}10$ OOD setting, and closes the gap to ANYCSP on MaxCut instances.
\end{itemize}

While neural methods offer promising heuristics for constraint satisfaction, the overlay of an LNS framework has clearly provided many avenues for improvement. To this end, future work is needed to fully cross-pollinate from the LNS literature to iterative neural approaches in order to reliably outperform both historical LNS and neural methods across a variety of problems. 
Finally, there is further potential to extend the LNS paradigm to neural Diffusion models~\cite{shi2024simplified,cardei2025constrained,hamu2025accelerated} that bear a striking resemblance to LNS methods and may offer more constrained and controllable diffusion in a range of generative AI applications.

\subsubsection{\ackname}
\begin{credits}
We thank the anonymous reviewers for their insightful feedback. This work was supported by the Institute of Information \& Communications Technology Planning \& Evaluation (IITP) grant funded by the Korean Government (MSIT) (No.~RS-2024-00457882, National AI Research Lab Project).
\end{credits}

\bibliographystyle{splncs04}
\bibliography{bibliography}

@article{popescu2022overview,
  title={An overview of machine learning techniques in constraint solving},
  author={Popescu, Andrei and Polat-Erdeniz, Seda and Felfernig, Alexander and Uta, Mathias and Atas, M{\"u}sl{\"u}m and Le, Viet-Man and Pilsl, Klaus and Enzelsberger, Martin and Tran, Thi Ngoc Trang},
  journal={Journal of Intelligent Information Systems},
  volume={58},
  number={1},
  pages={91--118},
  year={2022},
  publisher={Springer}
}

@article{bengio2021machine,
  title={Machine learning for combinatorial optimization: a methodological tour d’horizon},
  author={Bengio, Yoshua and Lodi, Andrea and Prouvost, Antoine},
  journal={European Journal of Operational Research},
  volume={290},
  number={2},
  pages={405--421},
  year={2021},
  publisher={Elsevier}
}

@inproceedings{
consformer,
title={Self-Supervised Transformers as Iterative Solution Improvers for Constraint Satisfaction},
author={Yudong Xu and Wenhao Li and Scott Sanner and Elias Boutros Khalil},
booktitle={Forty-second International Conference on Machine Learning},
year={2025},
url={https://openreview.net/forum?id=IQN6ID0snT}
}

@inproceedings{shaw1998using,
  title={Using constraint programming and local search methods to solve vehicle routing problems},
  author={Shaw, Paul},
  booktitle={International conference on principles and practice of constraint programming},
  pages={417--431},
  year={1998},
  organization={Springer}
}

@article{mara2022survey,
  title={A survey of adaptive large neighborhood search algorithms and applications},
  author={Mara, Setyo Tri Windras and Norcahyo, Rachmadi and Jodiawan, Panca and Lusiantoro, Luluk and Rifai, Achmad Pratama},
  journal={Computers \& Operations Research},
  volume={146},
  pages={105903},
  year={2022},
  publisher={Elsevier}
}

@incollection{lnsbook,
  title={Large neighborhood search},
  author={Pisinger, David and Ropke, Stefan},
  booktitle={Handbook of metaheuristics},
  pages={99--127},
  year={2018},
  publisher={Springer}
}

@inproceedings{
wang2025remasking,
title={Remasking Discrete Diffusion Models with Inference-Time Scaling},
author={Guanghan Wang and Yair Schiff and Subham Sekhar Sahoo and Volodymyr Kuleshov},
booktitle={The Thirty-ninth Annual Conference on Neural Information Processing Systems},
year={2025},
url={https://openreview.net/forum?id=IJryQAOy0p}
}

@InProceedings{Chang_2022_CVPR,
    author    = {Chang, Huiwen and Zhang, Han and Jiang, Lu and Liu, Ce and Freeman, William T.},
    title     = {MaskGIT: Masked Generative Image Transformer},
    booktitle = {Proceedings of the IEEE/CVF Conference on Computer Vision and Pattern Recognition (CVPR)},
    year      = {2022},
    pages     = {11315-11325}
}

@article{shi2024simplified,
  title={Simplified and generalized masked diffusion for discrete data},
  author={Shi, Jiaxin and Han, Kehang and Wang, Zhe and Doucet, Arnaud and Titsias, Michalis},
  journal={Advances in neural information processing systems},
  volume={37},
  pages={103131--103167},
  year={2024}
}

@inproceedings{
hamu2025accelerated,
title={Accelerated Sampling from Masked Diffusion Models via Entropy Bounded Unmasking},
author={Heli Ben-Hamu and Itai Gat and Daniel Severo and Niklas Nolte and Brian Karrer},
booktitle={The Thirty-ninth Annual Conference on Neural Information Processing Systems},
year={2025},
url={https://openreview.net/forum?id=WBcBhT1NKO}
}

@inproceedings{
yang2023learning,
title={Learning to Solve Constraint Satisfaction Problems with Recurrent Transformer},
author={Zhun Yang and Adam Ishay and Joohyung Lee},
booktitle={The Eleventh International Conference on Learning Representations },
year={2023},
url={https://openreview.net/forum?id=udNhDCr2KQe}
}

@inproceedings{
fan2025looped,
title={Looped Transformers for Length Generalization},
author={Ying Fan and Yilun Du and Kannan Ramchandran and Kangwook Lee},
booktitle={The Thirteenth International Conference on Learning Representations},
year={2025},
url={https://openreview.net/forum?id=2edigk8yoU}
}

@inproceedings{
kim2025train,
title={Train for the Worst, Plan for the Best: Understanding Token Ordering in Masked Diffusions},
author={Jaeyeon Kim and Kulin Shah and Vasilis Kontonis and Sham M. Kakade and Sitan Chen},
booktitle={Forty-second International Conference on Machine Learning},
year={2025},
url={https://openreview.net/forum?id=DjJmre5IkP}
}

@inproceedings{anycsp,
author = {T\"{o}nshoff, Jan and Kisin, Berke and Lindner, Jakob and Grohe, Martin},
title = {One model, any CSP: graph neural networks as fast global search heuristics for constraint satisfaction},
year = {2023},
isbn = {978-1-956792-03-4},
doi = {10.24963/ijcai.2023/476},
booktitle = {Proceedings of the Thirty-Second International Joint Conference on Artificial Intelligence},
articleno = {476},
numpages = {9},
location = {Macao, P.R.China},
series = {IJCAI '23}
}

@inproceedings{satnet,
  title={Satnet: Bridging deep learning and logical reasoning using a differentiable satisfiability solver},
  author={Wang, Po-Wei and Donti, Priya and Wilder, Bryan and Kolter, Zico},
  booktitle={International Conference on Machine Learning},
  pages={6545--6554},
  year={2019},
  organization={PMLR}
}

@article{rrn,
  title={Recurrent relational networks},
  author={Palm, Rasmus and Paquet, Ulrich and Winther, Ole},
  journal={Advances in neural information processing systems},
  volume={31},
  year={2018}
}

@InProceedings{ired,
    author    = {Du, Yilun and Mao, Jiayuan and Tenenbaum, Joshua B.},
    title     = {Learning Iterative Reasoning through Energy Diffusion},
    booktitle = {International Conference on Machine Learning (ICML)},
    year      = {2024}
}

@article{HOTTUNG2022103786,
title = {Neural large neighborhood search for routing problems},
journal = {Artificial Intelligence},
volume = {313},
pages = {103786},
year = {2022},
issn = {0004-3702},
author = {André Hottung and Kevin Tierney}
}

@incollection{hottung2020neural,
  title={Neural Large Neighborhood Search for the Capacitated Vehicle Routing Problem},
  author={Hottung, Andr{\'e} and Tierney, Kevin},
  booktitle={ECAI 2020},
  pages={443--450},
  year={2020},
  publisher={IOS Press}
}

@article{
hottung2025neural,
title={Neural Deconstruction Search for Vehicle Routing Problems},
author={Andr{\'e} Hottung and Paula Wong-Chung and Kevin Tierney},
journal={Transactions on Machine Learning Research},
issn={2835-8856},
year={2025},
url={https://openreview.net/forum?id=bCmEP1Ltwq},
note={}
}

@article{song2020general,
  title={A general large neighborhood search framework for solving integer linear programs},
  author={Song, Jialin and Yue, Yisong and Dilkina, Bistra and others},
  journal={Advances in Neural Information Processing Systems},
  volume={33},
  pages={20012--20023},
  year={2020}
}

@inproceedings{
wu2021learning,
title={Learning Large Neighborhood Search Policy for Integer Programming},
author={Yaoxin Wu and Wen Song and Zhiguang Cao and Jie Zhang},
booktitle={Advances in Neural Information Processing Systems},
editor={A. Beygelzimer and Y. Dauphin and P. Liang and J. Wortman Vaughan},
year={2021},
url={https://openreview.net/forum?id=IaM7U4J-w3c}
}

@article{sonnerat2021learning,
  title={Learning a large neighborhood search algorithm for mixed integer programs},
  author={Sonnerat, Nicolas and Wang, Pengming and Ktena, Ira and Bartunov, Sergey and Nair, Vinod},
  journal={arXiv preprint arXiv:2107.10201},
  year={2021}
}

@inproceedings{huang2023searching,
  title={Searching large neighborhoods for integer linear programs with contrastive learning},
  author={Huang, Taoan and Ferber, Aaron M and Tian, Yuandong and Dilkina, Bistra and Steiner, Benoit},
  booktitle={International conference on machine learning},
  pages={13869--13890},
  year={2023},
  organization={PMLR}
}

@article{zhou2023learning,
  title={Learning large neighborhood search for vehicle routing in airport ground handling},
  author={Zhou, Jianan and Wu, Yaoxin and Cao, Zhiguang and Song, Wen and Zhang, Jie and Chen, Zhenghua},
  journal={IEEE Transactions on knowledge and data engineering},
  volume={35},
  number={9},
  pages={9769--9782},
  year={2023},
  publisher={IEEE}
}

@article{feng2025spl,
  title={SPL-LNS: Sampling-Enhanced Large Neighborhood Search for Solving Integer Linear Programs},
  author={Feng, Shengyu and Sun, Zhiqing and Yang, Yiming},
  journal={arXiv preprint arXiv:2508.16171},
  year={2025}
}

@article{cappart2025combining,
  title={Combining Constraint Programming and Machine Learning: From Current Progress to Future Opportunities},
  author={Cappart, Quentin and Guns, Tias and Lombardi, Michele and Pesant, Gilles and Tsouros, Dimos},
  journal={Journal of Artificial Intelligence Research},
  volume={84},
  year={2025}
}

@inproceedings{
feng2025comprehensive,
title={Frontier{CO}: Real-World and Large-Scale Evaluation of Machine Learning Solvers for Combinatorial Optimization},
author={Shengyu Feng and Weiwei Sun and Shanda Li and Ameet Talwalkar and Yiming Yang},
booktitle={The Fourteenth International Conference on Learning Representations},
year={2026},
url={https://openreview.net/forum?id=BVprkacwFY}
}

@article{falkner2022large,
  title={Large neighborhood search based on neural construction heuristics},
  author={Falkner, Jonas K and Thyssens, Daniela and Schmidt-Thieme, Lars},
  journal={arXiv preprint arXiv:2205.00772},
  year={2022}
}

@inproceedings{
dehghani2018universal,
title={Universal Transformers},
author={Mostafa Dehghani and Stephan Gouws and Oriol Vinyals and Jakob Uszkoreit and Lukasz Kaiser},
booktitle={International Conference on Learning Representations},
year={2019},
url={https://openreview.net/forum?id=HyzdRiR9Y7},
}

@article{jolicoeur2025less,
  title={Less is more: Recursive reasoning with tiny networks},
  author={Jolicoeur-Martineau, Alexia},
  journal={arXiv preprint arXiv:2510.04871},
  year={2025}
}

@inproceedings{sanokowski2024diffusion,
  title={A Diffusion Model Framework for Unsupervised Neural Combinatorial Optimization},
  author={Sanokowski, Sebastian and Hochreiter, Sepp and Lehner, Sebastian},
  booktitle={International Conference on Machine Learning},
  pages={43346--43367},
  year={2024},
  organization={PMLR}
}

@article{sun2023difusco,
  title={Difusco: Graph-based diffusion solvers for combinatorial optimization},
  author={Sun, Zhiqing and Yang, Yiming},
  journal={Advances in neural information processing systems},
  volume={36},
  pages={3706--3731},
  year={2023}
}

@misc{ye2003gset,
  title={The gset dataset},
  author={Ye, Yinyu},
  year={2003},
  publisher={Stanford}
}

@article{ropke2006adaptive,
  title={An adaptive large neighborhood search heuristic for the pickup and delivery problem with time windows},
  author={Ropke, Stefan and Pisinger, David},
  journal={Transportation science},
  volume={40},
  number={4},
  pages={455--472},
  year={2006},
  publisher={Informs}
}

@misc{cpsat,
  title = {{CP-SAT}},
  version = { v9.11 },
  author = {Laurent Perron and Frédéric Didier},
  organization = {Google},
  url = {https://developers.google.com/optimization/cp/cp\_solver/},
  date = { 2024-05-07 }
}

@article{ecord,
  title={Learning to solve combinatorial graph partitioning problems via efficient exploration},
  author={Barrett, Thomas D and Parsonson, Christopher WF and Laterre, Alexandre},
  journal={arXiv preprint arXiv:2205.14105},
  year={2022}
}

@inproceedings{ECODQN,
  title={Exploratory combinatorial optimization with reinforcement learning},
  author={Barrett, Thomas and Clements, William and Foerster, Jakob and Lvovsky, Alex},
  booktitle={Proceedings of the AAAI conference on artificial intelligence},
  volume={34},
  number={04},
  pages={3243--3250},
  year={2020}
}

@article{runcsp,
  title={Graph neural networks for maximum constraint satisfaction},
  author={Toenshoff, Jan and Ritzert, Martin and Wolf, Hinrikus and Grohe, Martin},
  journal={Frontiers in artificial intelligence},
  volume={3},
  pages={580607},
  year={2021},
  publisher={Frontiers Media SA}
}

@article{sdp,
  title={Improved approximation algorithms for maximum cut and satisfiability problems using semidefinite programming},
  author={Goemans, Michel X and Williamson, David P},
  journal={Journal of the ACM (JACM)},
  volume={42},
  number={6},
  pages={1115--1145},
  year={1995},
  publisher={ACM New York, NY, USA}
}

@article{cardei2025constrained,
  title={Constrained Discrete Diffusion},
  author={Cardei, Michael and Christopher, Jacob K and Hartvigsen, Thomas and Bartoldson, Brian R and Kailkhura, Bhavya and Fioretto, Ferdinando},
  journal={arXiv preprint arXiv:2503.09790},
  year={2025}
}

@article{qiu2022dimes,
  title={Dimes: A differentiable meta solver for combinatorial optimization problems},
  author={Qiu, Ruizhong and Sun, Zhiqing and Yang, Yiming},
  journal={Advances in Neural Information Processing Systems},
  volume={35},
  pages={25531--25546},
  year={2022}
}

\end{document}